\documentclass{article}

\usepackage{PRIMEarxiv}

\usepackage[utf8]{inputenc} 
\usepackage[T1]{fontenc}    
\usepackage{hyperref}       
\usepackage{url}            
\usepackage{booktabs}       
\usepackage{amsfonts}       
\usepackage{nicefrac}       
\usepackage{microtype}      
\usepackage{lipsum}
\usepackage{fancyhdr}       
\usepackage{graphicx}       
\graphicspath{{media/}}     

\usepackage{amsmath,amssymb,amsfonts}
\usepackage{algorithmic}
\usepackage{graphicx}
\usepackage{textcomp}
\def\S{\mathbf{S}}

\def\y{\mathbf{y}}
\def\L\mathcal{L}

\def\V{\mathbf{V}}
\def\F{\mathbf{F}}

\def\x{\mathbf{x}}

\def\l{\mathbf{l}}

\def\L{\mathcal{L}}
\def\E{\mathbf{E}}

\def\y{\mathbf{y}}

\def\x{\mathbf{x}}

\def\F{\mathbf{F}}

\title{Expert Uncertainty and Severity Aware Chest X-Ray Classification by Multi-Relationship Graph Learning
\thanks{\textit{\underline{Citation}}: 
\textbf{Authors. Title. Pages.... DOI:000000/11111.}} 
}

\author{
  Mengliang Zhang, Xinyue Hu\\
  Department of Computer Science and Engineering \\
  University of Texas at Arlington \\
  Arlington, TX 76010, USA\\
  \texttt{\{mxz3935, xxh4034\}@mavs.uta.edu} \\
   \And
  Lin Gu, Tatsuya Harada\\
  RIKEN, University of Tokyo \\
  Tokyo, Japan\\
  \texttt{lin.gu@rikrn.jp, harada@mi.t.u-tokyo.ac.jp} \\
   \And
  Liangchen Liu, Ronald M.Summers\\
  National Institute of Health Clinical Center \\
  Bethesda, MD 20892 USA\\
  \texttt{liangchen.liu@nih.gov, rsummers@mail.cc.nih.gov} \\
   \And
  Kazuma Kobayashi\\
  National Cancer Center Research Institute \\
  Tokyo, Japan\\
  \texttt{kazumkob@ncc.go.jp} \\
   \And
  Yingying Zhu* \\
  Department of Computer Science and Engineering \\
  University of Texas at Arlington \\
  Arlington, TX 76010, USA\\
  \texttt{yingying.zhu@uta.edu} \\
}

\begin{document}
\maketitle

\begin{abstract}
    Patients undergoing chest X-rays (CXR) often endure multiple lung diseases. When evaluating a patient's condition, due to the complex pathologies, subtle texture changes of different lung lesions in images, and patient condition differences, radiologists may make uncertain even when they have experienced long-term clinical training and professional guidance, which makes much noise in extracting disease labels based on CXR reports. In this paper, we re-extract disease labels from CXR reports to make them more realistic by considering disease severity and uncertainty in classification. Our contributions are as follows: 1. We re-extracted the disease labels with severity and uncertainty by a rule-based approach with keywords discussed with clinical experts. 2. To further improve the explainability of chest X-ray diagnosis, we designed a multi-relationship graph learning method with an expert uncertainty-aware loss function. 3. Our multi-relationship graph learning method can also interpret the disease classification results. Our experimental results show that models considering disease severity and uncertainty outperform previous state-of-the-art methods.
\end{abstract}


\section{Introduction}
    Chest radiography is a routinely used imaging method to identify acute and chronic cardiopulmonary conditions and to assist in related medical workups. Although recent work shows impressive performance with large-scale text-mined labels on CXR image classification using datasets including ChestXray14, CheXpert, and MIMIC-CXR \cite{x.wangchestxray8hospitalscalechest2017, irvinchexpertlargechest2019, johnsonmimiccxrdeidentifiedpublicly2019}. There are several limitations in the current text-mined labeled CXR image dataset.

    Existing datasets have not thoroughly investigated the findings and impressions sections. The findings section contains information directly observed from the image, while the impression section is derived from a comprehensive diagnosis with external data like patients' history. Notably, the MIMIC-CXR dataset only focused on extracting labels from the impression section, disregarding the findings section. 
    
\begin{figure*}[h]
    \centering
    \includegraphics[width=0.8\linewidth] {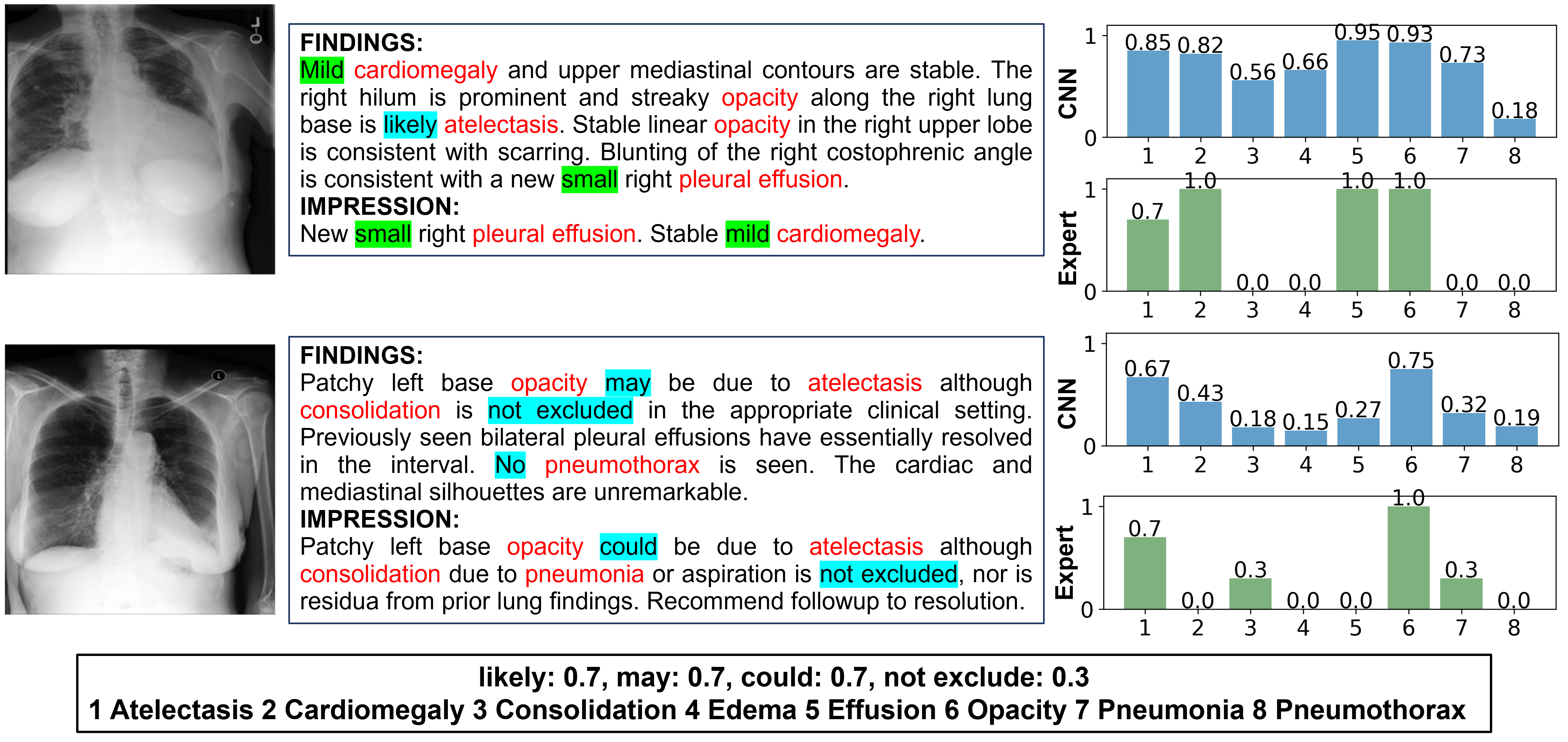}
    \caption{Extracted labels and predictions comparison between MIMIC-CXR and our method. Disease name, level, and uncertainty are highlighted in red, green, and blue. MIMIC-CXR labels are 0-1 encoded. Our labels are soft, representing different uncertainty levels through distinct label values.}
    \label{fig: fig1}
\end{figure*}

    In the MIMIC-CXR report,  diseases are frequently accompanied by their severity (e.g., 'small effusion' and 'mild cardiomegaly' in Figure \ref{fig: fig1}, more statistics can be found in Table \ref{tab: disease data}), carrying much clinical information for diagnose. To the best of our knowledge, no previous studies have addressed the issue of reported disease severity.

    Besides, current CXR image classification research has not considered the uncertainty information during model training. Radiologists implicitly indicate varying levels of uncertainty when writing clinical notes, using terms such as "likely", "not excluded", and "maybe" to indicate the degree of uncertainty in their diagnosis, as demonstrated in Figure \ref{fig: fig1}. Our analysis of the MIMIC-CXR report shows that certain disease descriptions contain numerous uncertain descriptions. However, less research has focused on addressing the issue of disease uncertainty. For instance, CheXpert \cite{irvinchexpertlargechest2019} treats the uncertainty label as either a negative or positive class or trains a clean label separately and then predicts the uncertainty of the label on the real label. Typically, CXR disease classification approaches \cite{x.wangchestxray8hospitalscalechest2017}, \cite{johnsonmimiccxrdeidentifiedpublicly2019}, \cite{x.ouyanglearninghierarchicalattention2021}, \cite{b.chenlabelcooccurrencelearning2020} treat uncertain labels as negative classes. Yang \cite{yanglearnbeuncertain2019} accounts for the uncertainty of diseases and investigates the relationship between uncertainty in diagnostic CXR radiology reports and uncertainty estimation of corresponding DNN models using Bayesian approaches. Moreover, the co-occurrence and dependence between diseases also contribute to uncertainty. CheXGCN \cite{b.chenlabelcooccurrencelearning2020} proposes a label co-occurrence learning module to generalize the relationship between pathologies into a set of classifier scores by introducing the word embedding of pathologies and multi-layer graph information propagation. This method can adaptively recalibrate multi-label outputs during end-to-end training using these scores. None of these methods reasonably quantifies the uncertainty to fit the actual situation.
    
    Furthermore, current CXR image classification models have yet to delve into the underlying reasons for predicting various diseases. Interpreting medical imaging is crucial as it aids radiologists in verifying the accuracy of deep learning models and building trust in AI systems. Previous studies \cite{irvinchexpertlargechest2019}, \cite{b.chenlabelcooccurrencelearning2020}, \cite{zhoucontrastattentivethoracicdisease2021}, \cite{zhaocrosschestgraph2021} have mainly employed the grad-cam \cite{r.r.selvarajugradcamvisualexplanations22} method to generate heat maps that localize the disease, thereby explaining the model's decisions. However, this method only provides the CNN model's degree of attention to various image positions and lacks a more in-depth explanation.

    In this study, we propose a threefold approach to address the current limitations of CXR image classification. Firstly, we construct an expert uncertainty and disease severity awareness dataset by extracting uncertainty and severity keywords from CXR clinical reports. We collaborate with radiologists to determine the potential probabilities for different uncertainty keywords. Based on this information, we implement a baseline model integrating expert-level uncertainties into CXR classification models.
    
    Secondly, we propose an anatomical structure-aware multi-relationship graph that enhances the interpretability of CXR classification models. We treat each anatomical region as a graph node and establish spatial and semantic relationships to connect them. The spatial relationship is based on the spatial location of different anatomical regions, while the semantic relationship is based on two CXR knowledge graphs we constructed. We evaluate the interpretability of this multi-relationship graph by measuring the contributions of graph nodes and edges in predicting different disease labels.

    Finally, we train our proposed anatomical region multi-relationship graph using uncertainty-aware and conventional labels from the MIMIC CXR dataset. We compare our approach with several state-of-the-art benchmark methods and demonstrate that incorporating expert-level uncertainty labels significantly improves classification accuracy and interpretability compared to conventional labels. Our contributions can be summarized as follows:

\begin{itemize}
    \item  We construct a large-scale chest X-ray image classification dataset that is expertly annotated with uncertainty and disease severity information based on clinical notes. This dataset enables the definition of different levels of uncertainties to be used as labels in training models. 

    \item We propose a simple but effective loss function considering the expert-level uncertainty labels during training.

    \item We design an anatomical structure-aware multi-relationship graph learning method that can provide a reasoning process for predicting different labels on graph nodes and edges, thereby improving interpretability. Our proposed method consistently outperforms benchmark methods regarding both performance and interpretability. We plan to make our dataset and source code publicly available upon publication of our work.
\end{itemize}

\section{Related Work}
\subsection{CXR disease classification using text-mined labels}
    Prior research has focused on constructing CXR datasets with clinical notes extracted labels, including Chest X-Ray14 \cite{x.wangchestxray8hospitalscalechest2017}, ChestXpert \cite{irvinchexpertlargechest2019}, and MIMIC-CXR \cite{johnsonmimiccxrdeidentifiedpublicly2019}. Studies ~\cite{x.wangchestxray8hospitalscalechest2017}, \cite{rajpurkarchexnetradiologistlevelpneumonia2017}, \cite{z.lithoracicdiseaseidentification2018}, \cite{liualignattendlocate2019} have utilized DenseNet and ResNet models for classifying abnormalities on CXR images with promising results. Some works \cite{g.zhaocontralaterallyenhancednetworks2021}, \cite{zhoucontrastattentivethoracicdisease2021} have proposed contrastive learning on the left and right lung regions to improve the localization of abnormal regions. Ouyang et al. \cite{x.ouyanglearninghierarchicalattention2021} proposed a novel hierarchical attention framework comprised of activation- and gradient-based attention mechanisms to solve the CXR image classification and abnormality localization problem to improve the performance of CXR image diagnosis. Chen et al. \cite{b.chenlabelcooccurrencelearning2020} proposed GCN-based architecture to leverage label co-occurrence and interdependency information for improving the multi-label CXR image classification task.

    Several recent works on chest X-ray image analysis and clinical notes generation leveraged expert knowledge graph as prior \cite{liknowledgedrivenencoderetrieve2019}, \cite{zhoucontrastattentivethoracicdisease2021} to improve the CXR classification, localization, and clinical notes generation. 
    
    However, to our knowledge, prior work has yet to systematically consider disease label uncertainty and severity. Therefore, we propose constructing a new dataset that includes different levels of medical expert uncertainty and disease severity. In addition, we constructed a new anatomical and co-occurrence CXR-related disease knowledge graph and considered both spatial and semantic relationships to improve our model's intepretability.

\subsection{Classification with uncertain and noisy label}

    Many previous work has studied the label noisy and uncertainty problem \cite{wanglearningnoisylabels2021}, \cite{renlearningreweightexamples2018}, \cite{y.lilearningnoisylabels22}, \cite{huanguncertaintyawarelearninglabel2022}, \cite{xiasampleselectionuncertainty2021}, \cite{petersonhumanuncertaintymakes2019}, \cite{zhanggeneralizedcrossentropy2018}, \cite{lukasikdoeslabelsmoothing2020}. These methods can be categorized into: 

\begin{itemize}
    \item Sample re-weighting based approach \cite{f.malearningnoisylabels2022}, \cite{renlearningreweightexamples2018}, \cite{liuclassificationnoisylabels2016}, which estimated the weights of noisy labeled samples and optimized a sample re-weighted loss function 
    \cite{liuclassificationnoisylabels2016}, \cite{xiasampleselectionuncertainty2021}, \cite{hancoteachingrobusttraining2018}, \cite{jiangmentornetlearningdatadriven2018}. 

    \item One popular approach for mitigating the effects of label noise involves estimating the noise transition matrix \cite{hendrycksusingtrusteddata2018}, \cite{hanmaskingnewperspective2018}, \cite{yanglearnbeuncertain2019}, which quantifies the probability of clean labels being corrupted during the labeling process. 
    
    \item Some studies have explored using soft labels to smooth the model \cite{lukasikdoeslabelsmoothing2020}, \cite{petersonhumanuncertaintymakes2019}. 
    
    \item leverages human labeled uncertainties explicitly \cite{petersonhumanuncertaintymakes2019} and trains the classification model to predict the human uncertainty distribution, which achieves more robust classification results.
\end{itemize}

    Furthermore, the impact of expert annotation on label noise is often overlooked. In this study, we propose a method that identifies and utilizes uncertain labels from reports written by experts, enabling further advancement in medical image noisy label learning.

\begin{table*}[htbp]
\caption{Table of keywords used to extract disease labels.}
\label{tab: disease keywords}
\centering
\begin{tabular}{l|l}
    \hline
    \textbf{Disease Label}       & \textbf{Key Words}     \\
    \hline
    Atelectasis         & atelecta, collapse \\
    Blunting of costophrenic angle   & blunting   of costophrenic angle \\
    Calcification       & calcification  \\
    Cardiomegaly        & cardiomegaly, the  heart, heart  size, cardiac enlargement, cardiac size, \\
        & cardiac shadow, cardiac contour, cardiac silhouette, enlarged heart, \\
        & mediastinum, cardiomediastinum, contour, mediastinal configuration \\  
        & mediastinal silhouette, pericardial silhouette,\\
        & cardiac silhouette and vascularity \\
    Consolidation      
        & consolidat, consolidation      \\
    Edema & edema, heart failure, chf, pulmonary congestion,\\ 
        &indistinctness,  vascular prominence  \\
    Emphysema           & emphysema             \\
    Fracture            & fracture              \\
    Granuloma           & granuloma             \\
    Hernia              & hernia \\
    Lung  Opacity      & opaci, decreased  translucency, increased   density, airspace   disease, \\ 
        & air-space  disease, air   space disease, infiltrate\\    
        & infiltration, interstitial   marking, interstitial   pattern,  interstitial  lung, \\
        & reticular   pattern, reticular   marking\\    
        & reticulation, parenchymal   scarring, peribronchial   thickening, wall   thickening, scar \\    
    Pleural  Effusion  & pleural   fluid, effusion         \\
    Pleural  Thickening& pleural   thickening  \\
    Pneumonia           & pneumonia, infection, infectious   process, infectious         \\
    Pneumothorax        & pneumothorax, pneumothoraces     \\
    Scoliosis           & scoliosis             \\
    Tortuosity   of the thoracic aorta & tortuosity   of the thoracic aorta   \\
    Vascular   congestion    & vascular   congestion \\
    \hline
\end{tabular}
\end{table*}

\begin{table}[]
\caption{We combined the words about the disease level extracted from the reports into several categories. Each category has similar disease severity.}
\label{tab: disease level}
\centering
\setlength{\tabcolsep}{5.5mm}{
\begin{tabular}{l|l}
    \hline
    \textbf{Merged Level} & \textbf{Extracted Words}\\
    \hline
    Mild  & mild, small, trace, minor, minimal\\
    & subtle, mildly, minimally \\
    Moderate & moderate, moderately, mild to moderate\\
    Severe & severe, acute, massive\\
    & moderate to   severe, moderate to large \\
    \hline
\end{tabular}}
\end{table}

\section{Methodology}
\subsection{Classification with hard labels.}
    Denotes a set of labelled CXR image dataset: $\mathcal{X}=[\x_1, \ldots,\x_N]$ represent the extracted image features, $\mathcal{Y} = [\y_1,\cdots,\y_N]$ represent the ground truth class labels. The standard CXR classification will train a label prediction model  $p_{\theta}(\y_i = \l_c|\x_i)$ for all observed training samples, where $\l_c$ is the label for class $c=1,\cdots, C$, $\theta$ is model's parameter. The model is trained to minimize the cross entropy between the predicted label distribution and the ground truth label distribution as,
\begin{equation}
    \arg\min_{\theta} -\frac{1}{N}\sum_{i=1}^{N} \sum_{c=1}^{C} p(\y_i=\l_c|\x_i) \log p_{\theta}(\y_i =\l_c|\x_i)
    \label{eq1}
\end{equation}
    The current CXR image classification assumes that the underlying conditional data distribution p(y|x) is zero for every category $c$ apart from the assigned class. A rule-based text mining approach extracted these assigned class labels without considering the label uncertainty. By contrast, when we consider the network and human expert uncertainty as seen in Figure ~\ref{fig: fig1}, we can see cases in which this assumption violates expert uncertainties on different class labels.

\subsection{Expert uncertainty and severity aware chest X-ray dataset construction. }

\textbf{Extract disease labels from both the impression and finding section.}
    In general, clinical reports for chest X-rays consist of two sections: Findings are detailed descriptions of all kinds of observations in the image, including both normal and abnormal ones. Impression is a summary of observations, which usually only has one or two sentences.

    The current datasets for chest X-rays extract image labels from either the finding section or the impression section alone, as noted in \cite{johnsonmimiccxrdeidentifiedpublicly2019}. However, since both sections contain crucial information for accurate diagnosis, it is important to consider both.

    In medical reports, the same disease can be expressed differently or using different terminology. For example, "cardiomegaly" and "heart size enlarged" can both refer to an enlarged heart. In order to accurately extract the disease labels from the clinical reports, a keyword list for different diseases is necessary to ensure that all possible variations of the disease names are covered.

    Table \ref{tab: disease keywords} includes synonyms, abbreviations, and variations of the disease names to ensure that all possible expressions of the disease are included in the dataset. The keyword table was created in collaboration with experienced radiologists and medical experts, who provided their input on the different terminologies used in clinical reports.

\begin{table}[htbp]
\caption{Words with different degrees of uncertainty are quantified into different label probability and rank.}
\label{tab: assign uncertain words}
    \centering
    \setlength{\tabcolsep}{5mm}{
    \begin{tabular}{c|cl}
    \hline
    \textbf{Rank} & \textbf{Probability} & \textbf{ Uncertain keywords} \\
    \hline
    1 & 1.0 & positive, change in.\\
    2 & 0.7 & probable, likely, may,\\ && could, potential.\\
    3 & 0.5 & might, possible.\\ 
    4 & 0.3 & not exclude,\\ && cannot accurately assesses,\\ &&
    cannot assessed, cannot identified,\\ && cannot confirmed, difficult exclude.\\
    5 & 0.1 & not mentioned.\\ 
    6 & 0.0 & no. \\ 
    \hline
    \end{tabular}}
\end{table}

\vspace{0.3cm}
\textbf{Extracting core disease labels along with their respective severities.}
    Our dataset was created by extracting disease-related labels from both the finding and impression sections, as illustrated in Figure~\ref{fig: fig1}. Prior works, including CheXpert \cite{irvinchexpertlargechest2019} and MIMIC-CXR ~\cite{johnsonmimiccxrdeidentifiedpublicly2019}, only considered a limited number of disease labels, typically around 14, without accounting for severity and uncertainty levels. This resulted in many significant diseases, such as granuloma, being omitted from previous datasets. In order to address this gap, we extracted 18  diseases to encompass a more comprehensive range of potential diseases, resulting in 60 disease-related keywords in our dataset. 
    
    Furthermore, we classified certain diseases into subcategories based on severity levels, such as "mild Cardiomegaly" and "severe Cardiomegaly", to more precisely capture the severity of the disease associated with expert interpretation of CXR images. The detailed table of merged keywords for each severity level can be found in Table \ref{tab: disease level}. This approach allowed us to capture the severity level of diseases in a standardized manner, enabling us to construct an expert uncertainty and disease severity-aware dataset for CXR image classification.

\begin{table}[htbp]
\caption{Distribution of uncertain labels (probabilities) for different diseases. These data count the diseases in the report corresponding to the image of the PA view in MIMIC-CXR.}
\label{tab: disease data}
\centering
\begin{tabular}{l|cccc}
    \hline
    {\textbf{Disease}} & \textbf{1.0} & \textbf{0.7} & \textbf{0.5} & \textbf{0.3} \\
    \hline
    Atelectasis    & 10651        & 3507& 177 & 554 \\
    Blunting   of  costophrenic angle    & 907 & 6 & 8 & 25\\
    Calcification  & 2889& 87& 9 & 5 \\
    Cardiomegaly   & 6846& 69& 15& 10\\
    Consolidation  & 1936& 268 & 59& 374 \\
    Edema & 2325& 405 & 59& 86\\
    Emphysema      & 1481& 122 & 21& 5 \\
    Fracture       & 2602& 140 & 42& 10\\
    Granuloma      & 953 & 308 & 24& 3 \\
    Hernia& 946 & 132 & 9 & 4 \\
    Lung   Opacity & 14605        & 163 & 51& 563 \\
    Pleural   Effusion        & 8319& 1030& 366 & 240 \\
    Pleural   Thickening      & 1342& 381 & 27& 8 \\
    Pneumonia      & 4356& 2219& 248 & 1404\\
    Pneumothorax   & 1322& 44& 19& 20\\
    Scoliosis      & 1333& 17& 1 & 1 \\
    Tortuosity   of the thoracic aorta   & 737 & 0 & 0 & 0 \\
    Vascular   congestion     & 1752& 112 & 67& 0 \\
    \hline
\end{tabular}
\end{table}

\textbf{Extracting disease labels while taking into account the level of uncertainty associated with expert interpretations.}
    MIMIC-CXR and CheXpert ~\cite{irvinchexpertlargechest2019} extracted disease labels with uncertainties and assigned a 'u' label to indicate uncertainty. However, actual uncertainties can be expressed through different keywords, such as "not excluded", "maybe", and "likely". Unfortunately, these important uncertainty level indicators were not considered in these previous works. Through consultations with radiologists, we assigned different probabilities to different levels of uncertainties based on the corresponding keywords, as illustrated in Table~\ref{tab: assign uncertain words}. Furthermore, in addition to assigning expert-level probabilities to each label, we also ranked the labels according to their respective levels of expert uncertainties.  In Table ~\ref{tab: disease data}, we counted the uncertainty distribution of diseases in reports corresponding to the postero-anterior (PA) view in MIMIC-CXR. It can be seen that some diseases are often described with uncertainty in the report, which is precisely what previous studies have ignored.


\subsection{Expert uncertainty level aware classification.}

    After constructing the dataset with expert uncertainty aware labels, each label is assigned with a medical expert aware probability score as stated in Table~\ref{tab: assign uncertain words}. We minimize the cross entropy between the predicted label distribution and the expert label distribution as,
\begin{equation}
    \arg\min_{\theta} -\frac{1}{N}\sum_{i=1}^{N} \sum_{c=1}^{C} p_{ex}(\y_i=l_c|\x_i) \log p_{\theta}(\y_i =l_c|\x_i)
    \label{eq:expert_uncertain}
\end{equation}
    where $p_{ex} (\y_i=l_c|\x_i)$ is the extracted expert probability score on label $l_c$ for image $\x_i$. Our method stands out from previous techniques by incorporating evidence-based uncertainties and training a model to predict the actual uncertainty distributions associated with expert-labeled diseases. This contrasts previous methods that either relied on pre-training with clean models or distillation methods to assign soft labels.

\begin{figure*}[htbp]
    \centering
    \includegraphics[width=16cm]{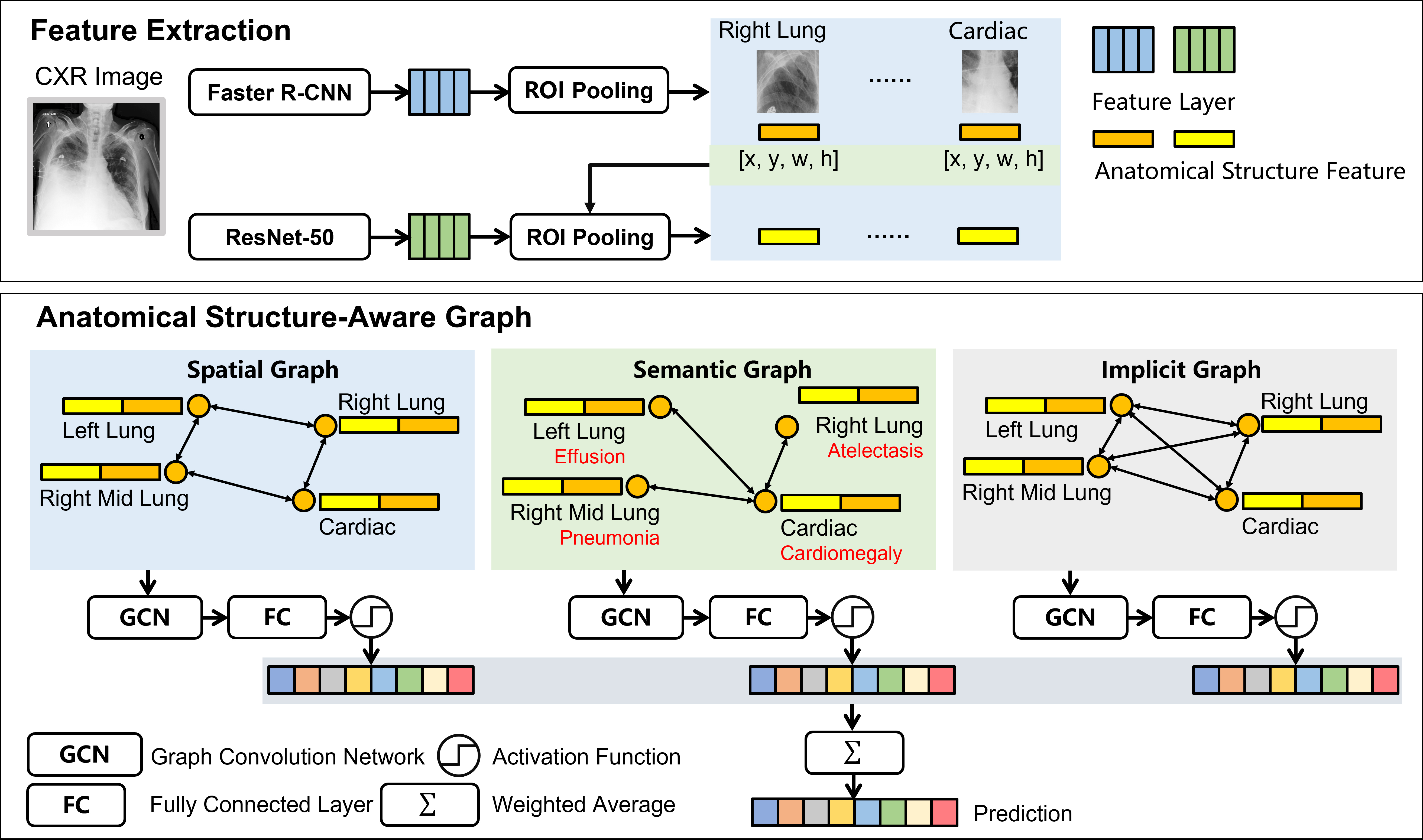}
    \caption{Architecture of our proposed method. The first part is a feature extraction module to obtain anatomical structure features from the detection and classification model by ROI pooling. The second part is the anatomical structure-aware graph module to construct three kinds of graphs. Finally, the predicted value is obtained by a weighted average of the predicted results of the three graphs.}
    \label{fig:architecture}
\end{figure*}

\subsection{Multi-Relationship Graph Learning}
    In order to model the predicted label probability, a function $g$ is constructed as $p_{\theta} (\y_i=\l_c|\x_x) =g_{\theta}(\x_i,\l_c)$. Since the radiologist reviews CXR images by first locating important anatomical regions and checking the abnormalities in different anatomical regions, to model how the radiologist makes the diagnosis, we proposed to construct a multi-relationship graph learning method to model the relationship between different anatomical regions and use it as $g$.
    We extract the features of different anatomical regions from chest X-ray images and consider each anatomical region as a graph node. We exploit Faster-RCNN \cite{renfasterrcnnrealtime2015} as the backbone of the anatomy detection model to train anatomical region detection ImaGenome \cite{wuchestimagenomedataset2021} dataset. 

    To select the relevant anatomical structures, we consulted with expert radiologists and identified key areas of the lungs and heart that are important for diagnosing diseases. We divided the lungs into three sub-regions, namely upper, middle, and lower, and marked significant areas such as the lung-diaphragmatic angle, which is the angle formed between the diaphragm and the chest wall. These marked areas were used as nodes in our graph network.

    Table \ref{tab: anatomy structure} provides 26 anatomical structures, including their names and descriptions. By incorporating these structures into our graph network, we aim to improve the interpretability of our model and provide insights into how different regions of the lungs are related to specific diseases.

\begin{table}[htbp]
\caption{We list some anatomical structures used in our graph construction. These anatomical structures are regarded as nodes in the graph.}
\label{tab: anatomy structure}
\centering
\begin{tabular}{l|l}
    \hline
    \textbf{Area} & \textbf{Anatomical Structures} \\   
    \hline
    Right Lung & Right lung, Right upper lung, Right mid lung, \\
        & Right lower lung, Hilar of right lung,\\
        & Apical of right lung, Right costophrenic sulcus,\\ 
        & Right hemidiaphragm. \\
    Left Lung  & Left lung, Left upper lung, Left mid lung,\\
        & Left lower lung, Hilar of left lung, \\
        & Apical of   left lung, Left costophrenic sulcus, \\
        & Left hemidiaphragm.   \\
    Cardiac    & Cardiac, Cavoatrial, Descending aorta,\\                   & Structure of carina. \\

    Others  & Main Bronchus, Right clavicle, Left   clavicle,\\  
        & Mediastinum, Aortic arch structure,\\
        & Superior vena cava structure.\\
    \hline   
\end{tabular}
\end{table}
    
    After obtaining the anatomy bounding boxes, we extracted the image region features using ROI pooling ~\cite{r.girshickfastrcnn07a}. 
    
    Table \ref{tab: anatomy structure detection} shows the dataset evaluation at the anatomical location (object) level. The F1 scores are calculated for relations extracted between objects and attributes from the 500 gold standard reports. Using the 1000 CXR images in the gold standard dataset, we also calculated the intersection over union (IoU) between the automatically extracted Bboxes and the validated and corrected Bboxes. Missing bounding boxes could be due to Bbox extraction failure or the anatomical location genuinely not being visible in the image (i.e., cut off or not in the field of view), which is not uncommon for the costophrenic angles and apical zones.

\begin{table}[htbp]
\caption{CXR image anatomical structure detection evaluation results}
\label{tab: anatomy structure detection}
\centering
\begin{tabular}{l|cccc}
\hline
\textbf{Bbox   name}   & \textbf{F1 score} & \textbf{Bbox IoU}  & \textbf{Missing Bbox} \\
\hline 
Left lung   & 0.933  & 0.976  & 0.03\%\\
Right lung  & 0.937  & 0.983   & 0.04\%\\
Cardiac   silhouette  & 0.966  & 0.967  & 0.01\%\\
Left lower lung   zone& 0.932  & 0.955  & 2.36\%\\
Right lower lung   zone    & 0.902  & 0.968  & 2.27\%\\
Right hilar   structures   & 0.934  & 0.976  & 1.91\%\\
Left hilar   structures    & 0.944  & 0.971  & 2.28\%\\
Upper mediastinum& 0.94   & 0.994  & 0.12\%\\
Left costophrenic   angle  & 0.908  & 0.929  & 0.63\%\\
Right costophrenic angle & 0.918  & 0.944  & 0.39\%\\
Left mid lung   zone  & 0.94   & 0.967  & 2.79\%\\
Right mid lung   zone & 0.83   & 0.968  & 2.31\%\\
Aortic arch & 0.965  & 0.991  & 0.62\%\\
Right upper lung zone    & 0.873  & 0.972  & 0.04\%\\
Reft upper lung zone& 0.811  & 0.968  & 0.22\%\\
Right hemidiaphragm & 0.947  & 0.955  & 0.15\%\\
Right clavicle   & 0.615  & 0.986  & 0.50\%\\
Left clavicle    & 0.642  & 0.983  & 0.51\%\\
Left hemidiaphragm  & 0.93   & 0.944  & 0.14\%\\
Right apical zone& 0.852  & 0.969  & 1.99\%\\
Trachea& 0.983  & 0.995  & 0.24\%\\
Left apical zone & 0.938  & 0.963  & 2.40\%\\
Carina& 0.975  & 0.994  & 1.47\%\\
Right atrium& 0.963  & 0.979  & 0.18\% \\  
\hline
\end{tabular}
\end{table}

\begin{figure}[htbp]
    \centering
    \includegraphics[width=10cm]
    {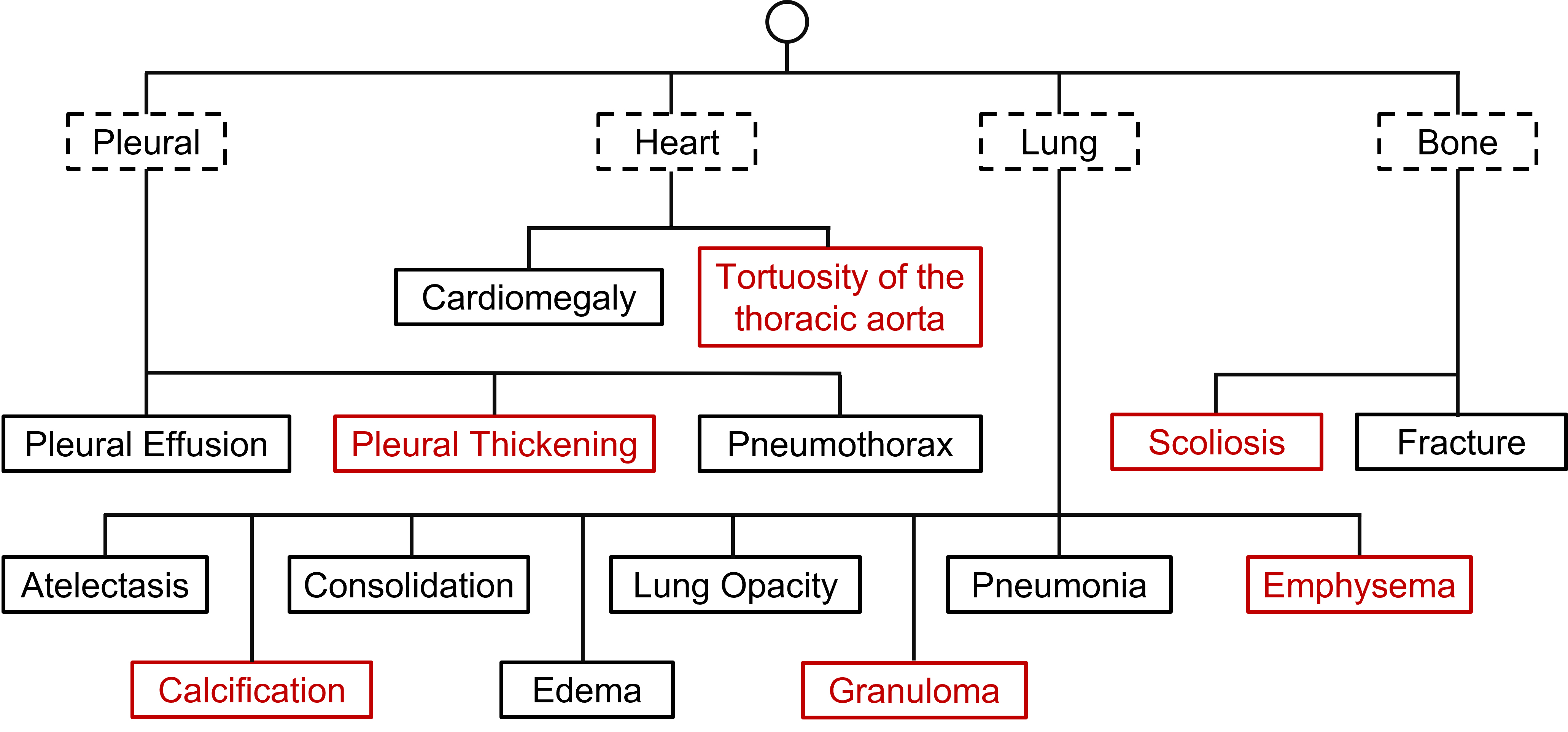}
    \caption{Anatomical structure and disease knowledge graph. The dotted line boxes are the primary anatomical structures, the solid line boxes are the findings, and the solid red line boxes are some of the disease types we added. The connection between the disease and the anatomical structure indicates that the disease mainly occurs on these anatomical structures.}
    \label{fig:anatomicalKG}
\end{figure}

Denote the location of all anatomical structures as $\mathcal{V} =[\V_1,\cdots, \V_m]$, and the feature of all anatomical structures as $\mathcal{F} =[\F_1,\cdots,\F_m]$, $m$ is the number of anatomical regions, we define a multi-relationship graph as $\mathcal{G}=\{\mathcal{V}, \mathcal{F},\mathcal{E}_{sp}, \mathcal{E}_{se}, \mathcal{E}_{im}\}$, where $\F_i$ represent the visual feature of the anatomical structure $i$, and $\V_i $ represent the location, height, and width of the structure $i$, $\mathcal{E}_{sp}$, $\mathcal{E}_{se}$ and $\mathcal{E}_{im}$ are the set of the spatial, semantic, and implicit edges respectively.

\textbf{Constructing spatial relationship edge.} $\mathcal{E}_{sp}$ represent the spatial relationship between these anatomical areas. We calculate the intersection of union (IOU) between two anatomical areas and set a threshold to decide whether they are connected. Given two anatomical region location $\V_{i}, \V_{j}$, the edge is calculated below:
\begin{equation}
    \E_{ij}= \begin{cases}1 & \text {if} \operatorname{IOU}\left(\V_i, \V_j\right) \geq \tau \\ 0 & \text {if} \operatorname{IOU}\left(\V_i, \V_j\right)<\tau\end{cases}
    \label{eq: IOU}
\end{equation}
    where $\tau$ is the threshold and $\operatorname{IOU}$ means IOU calculation. In our method, this threshold is set to 0.5. 
    
\vspace{0.3cm}    
\textbf{Constructing semantic relationship edge.} 
    The semantic edge $\mathcal{E}_{se})$ is constructed by leveraging the disease knowledge graph; see Figure \ref{fig:anatomicalKG}. Zhang et al. \cite{zhangwhenradiologyreport2020} describe the grouping of diseases in the constructed graph. Inspired by this, we also established disease groupings to describe the relationship between anatomical structures and diseases. Each node corresponding to an anatomical structure may have several kinds of disease. 
    
\begin{figure}[htbp]
    \centering
    \includegraphics[width=8.5cm]{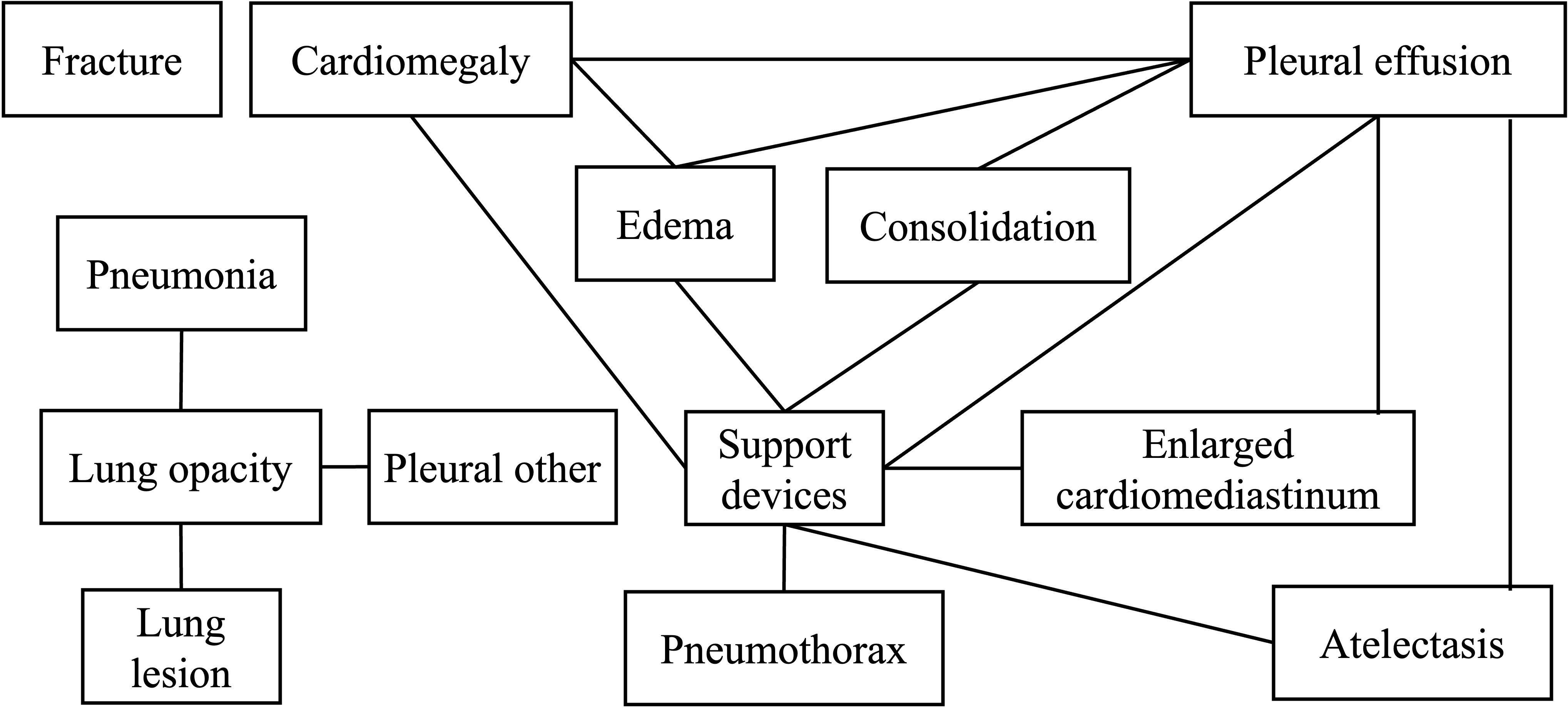}
    \caption{Disease co-occurrence relationship. The solid line boxes represent the diseases, and the connecting line between the two diseases has co-occurrences.}
    \label{fig:coKG}
\end{figure}

    We leverage Grad-CAM on the graph network \cite{popeexplainabilitymethodsgraph2019} to find every node's top-1 and top-2 predicted disease labels. We statistic the disease labels' co-occurrence matrix in the dataset. If two findings usually happen together, we regard these two findings as having a cooccurrence relationship; see Figure \ref{fig:coKG}. For top-1, top-2 diseases occur in one anatomical area; we construct the adjacent matrix $\mathbf{A}_{1}$ and  $\mathbf{A}_{2} \in \mathbb{R}^{K \times K}$ by the relationship mentioned above, where $K$ denotes the node number. Finally, the semantic adjacency matrix is the average of the corresponding elements of the two adjacency matrices.

\vspace{0.3cm}
\textbf{Constructing implicit relationship graph edge.} We consider a fully connected graph between all regions as $\mathcal{E}_{im}$, and this can potentially model the implicit relationships between the anatomical areas.

\vspace{0.3cm}
\textbf{Edge Feature Construction and Convolution.}
    We also compute edge features in our framework to model the node dependency relationship. Each edge $\text{edge}_{ij} $ between two nodes $ (\V_i, \V_j) \in \mathcal{V}$ has two attributes: 1) a representation of spatial structure relationship between nodes $\text{edge}_{ij}$, initialized with the normalized anatomical region center locations $[x_i, y_i, x_j, y_j]$ and updated in each layer of graph convolution; 2) a measure of dependency $e_{ij}$ between nodes $\V_i, \V_j$, initialized with the spatial, semantic or implicit relationship between two nodes and updated in each layer of graph convolution. 
    Such a design helps our framework discover human interpretable graph node and edge relationships contributing to network prediction and how their underlying interactions affect the final decision.

    Besides, we want to capture the relationship between nodes, finding edges related to specific nodes. To better capture inter-node relationships, we concatenate edge features with neighboring node features, denoted as $\mathcal{C}\left(e_{j k} \mathbf{W}_2 \mathbf{F}_j^l, \text{edge} _{jk}^{l}\right)$, and the equation becomes: 
\begin{equation}
    \mathbf{F}_{k}^{l+1}=\mathbf{W}_1 \mathbf{F}_k^l+\sum_{j \in \mathcal{N}(k)} \mathbf{W}_3 \mathcal{C}\left(e_{j k}^c \mathbf{W}_2 \mathbf{F}_j^l, \text{edge}_{j k}^l\right)
    \label{edge feature}
\end{equation}
    where edge feature $\text{edge}_{jk}^{l+1}=\mathbf{W}_4 \text{edge}_{jk}^{l}$, ${edge}_{jk}$ means the edge feature between node $\V_{j}$ and $\V_{k}$, $\mathbf{W}_3$ and $\mathbf{W}_4$ denoting one layer linear transformation for concatenated message feature. Since $e_{jk}^c$ is a trainable parameter by design, it helps learn the concept of inter-dependency as measured by the overall training objective.

    We compute graph convolution on the proposed three different relationship graphs to generate embedded image feature $\mathbf{F}_{sp}^L \in \mathbb{R}^{d\times K}, \mathbf{F}_{se}^L \in \mathbb{R}^{d\times K}, \mathbf{F}_{im}^L \in \mathbb{R}^{d \times K}$ at the last layer $L$. It is worth noting that the generated embedded feature includes both the node and edge features. Then, we apply an average pooling on the $K$ nodes to generate feature representation for each graph as $\mathbf{f}_{sp}^{*L} \in \mathbb{R}^{d\times 1}$, $\mathbf{f}_{se}^{*L} \in \mathbb{R}^{d\times 1}$, $\mathbf{f}_{im}^{*L} \in \mathbb{R}^{d\times 1}$.  
    We train three multi-layer perceptron (MLP) on these three graph features to predict the labels and output the final prediction as the weighted average of the predicted labels, 
\begin{equation}
    \hat{\mathbf{y}}_{k} = \alpha g_{sp} (f_{sp}^{*L}) + \beta g_{sp} (f_{se}^{*L}) + (1-\alpha - \beta) g_{im}(f_{im}^{*L})
    \label{eq: prediction}
\end{equation}
    where $g_{sp}, g_{se}, g_{im}$ are the MLP model for spatial, semantic, and implicit graphs; respectively, $\alpha$ and $\beta$ are hyper-parameters that control the proportion of predicted values for different graphs. We use the equation~\ref{eq:expert_uncertain} as the loss function to train this model.

\vspace{0.3cm}
\textbf{Visualization of important graph nodes and edge for predicted labels.} 
    We also visualize the important image regions and region-to-region contributions on predicted class labels to interpret which region in the image contributed to the classification results with our uncertainty labels. For each class, $c$ with the class prediction score $\hat{y}^c$, we compute gradients of $\hat{y}^c$ concerning the graph node and edge feature in the spatial, semantic, and implicit graph respectively as,
\begin{eqnarray}
    &&  \nonumber {\boldsymbol {B} }^c_{sp} = \frac{\partial \hat{y}^c}{\partial \mathbf{F}_{sp}^L}, {\boldsymbol {B} }^c_{se} = \frac{\partial \hat{y}^c}{\partial \mathbf{F}_{se}^L}, 
                   {\boldsymbol {B} }^c_{im} = \frac{\partial \hat{y}^c}{\partial \mathbf{F}_{im}^L} \\
    &&    
       c=1, \ldots, C
        \label{gradient}
\end{eqnarray}
    $\boldsymbol{B}^c_{sp} \in \Re^{d\times K }$ denotes the gradients matrix for spatial, semantic, and implicit graphs, respectively.
\begin{eqnarray}
    &&\nonumber    \mathbf{S}^c_{sp} = \boldsymbol{B}^c_{sp} \odot \mathbf{F}_{sp}^L,
         \mathbf{S}^c_{se} = \boldsymbol{B}^c_{se} \odot \mathbf{F}_{se}^L,
          \mathbf{S}^c_{im} = \boldsymbol{B}^c_{im} \odot \mathbf{F}_{im}^L
         \\ 
    &&   c=1, \ldots, C
        \label{importance}
\end{eqnarray}   
    The $\mathbf{S}^c_{sp}, \S^c_{se}, \S_{im}^c \in \Re^{d\times K}$ is the contribution of each node and edge in the spatial, semantic, and implicit graph under category $c$. We visualize the selected top nodes and top edges to interpret why this label is predicted in our model.
\begin{figure*}[htbp]
    \centering
    \includegraphics[width=16cm]
    {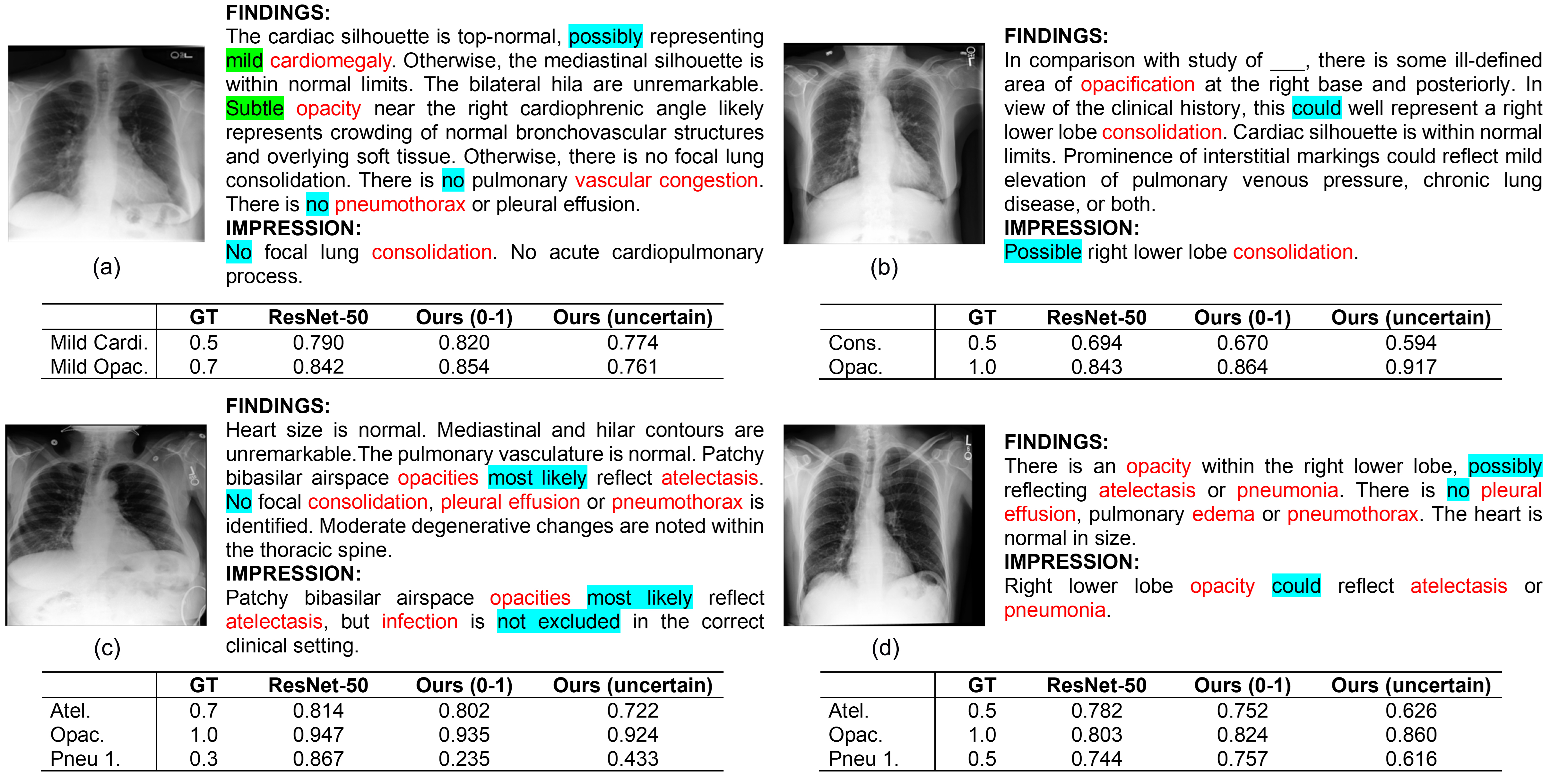}
    \caption{Experiment results comparison between several methods. The methods denoted by (0-1) following their names are trained on 0-1 labels, while those denoted by (uncertain) are trained on uncertain labels. Atel: Atelectasis, Card: Cardiomegaly, Cons: Consolidation, Opac: Opacity, Pneu 1: Pneumonia. Each example comprises a CXR image and the radiology report's corresponding findings and impressions section in the left and right panels. Findings, disease level, and uncertainty words are highlighted in red, green, and blue.}
    \label{fig: result}
\end{figure*}

\section{Experimental Results}
\subsection{Set up}
\textbf{Dataset.}
    We train our model on the MIMIC-CXR \cite{johnsonmimiccxrdeidentifiedpublicly2019} dataset. The MIMIC-CXR dataset consists of 377,110 CXR images of 65,379 patients, corresponding to 227,835 radiographic studies. Every study has a CXR report, which records the radiologist's diagnosis of the CXR image. We select images with PA view, vertical orientation, and corresponding reports as our dataset.

\vspace{0.3cm}
\textbf{Benchmark Methods.} 
We compare with several other methods:

\begin{itemize}
    \item ResNet-50 (0-1) trains on the 0-1 label with ResNet-50 backbone. In the labels we extracted, set all uncertain labels to 0 as hard labels. We also experimented with ResNet-50 on our extracted uncertain labels.

    \item ResNet-50 (uncertain) trains on the uncertain label with ResNet-50 backbone.

    \item Spatial Graph (0-1) uses the 0-1 label for training on the spatial graph we built. 

    \item Semantic Graph (0-1) uses the 0-1 label for training on the semantic graph we built.

    \item Implicit Graph (0-1) uses the 0-1 label for training on the implicit graph we built.

    \item Ours (0-1) uses the 0-1 label for training on the graph network we built.

    \item Ours (uncertain) uses the uncertain label for training on the graph network we built.
\end{itemize}

    The area-under-the-curve (AUC) score and top-k accuracy metrics are exploited to evaluate the performance of the listed methods. Table \ref{tab: experimental result} shows the average AUC for each method.

\begin{table}[ht]
    \caption{AUC and Top-k results. The methods denoted by an asterisk following their names are trained on uncertain labels, while those without an asterisk are trained on 0-1 labels. AUC and Top-k resuts. The methods denoted by (0-1) following their names are trained on 0-1 labels, while those denoted by (uncertain) are trained on uncertain labels.}
    \label{tab: experimental result}
    \centering
    \begin{tabular}{l|ccc}
    \hline
    \textbf{Method} & \textbf{Mean AUC (\%)} & \textbf{Top-5 (\%)} & \textbf{Top-10 (\%)}\\
    \hline
    {ResNet-50 (0-1)}  & {83.13}  & 74.65 & 85.44\\
    {ResNet-50 (uncertain)}  & {83.46}  & 74.94 & 85.82\\
    {Spatial Graph (0-1)} & {84.40} & 75.57 & 86.70\\
    {Semantic Graph (0-1)} & {84.52} & 75.73 & 86.90\\
    {Implicit Graph (0-1)} & {84.10} & 75.42 & 86.52\\
    {Ours (0-1)} & {85.27} & 76.36 & 87.32\\
    {Ours (uncertain)} & {86.09} & 77.18 & 88.49\\
    \hline
    \end{tabular}
\end{table}

\begin{figure}[htbp]
    \centering
    \includegraphics[width=7cm]{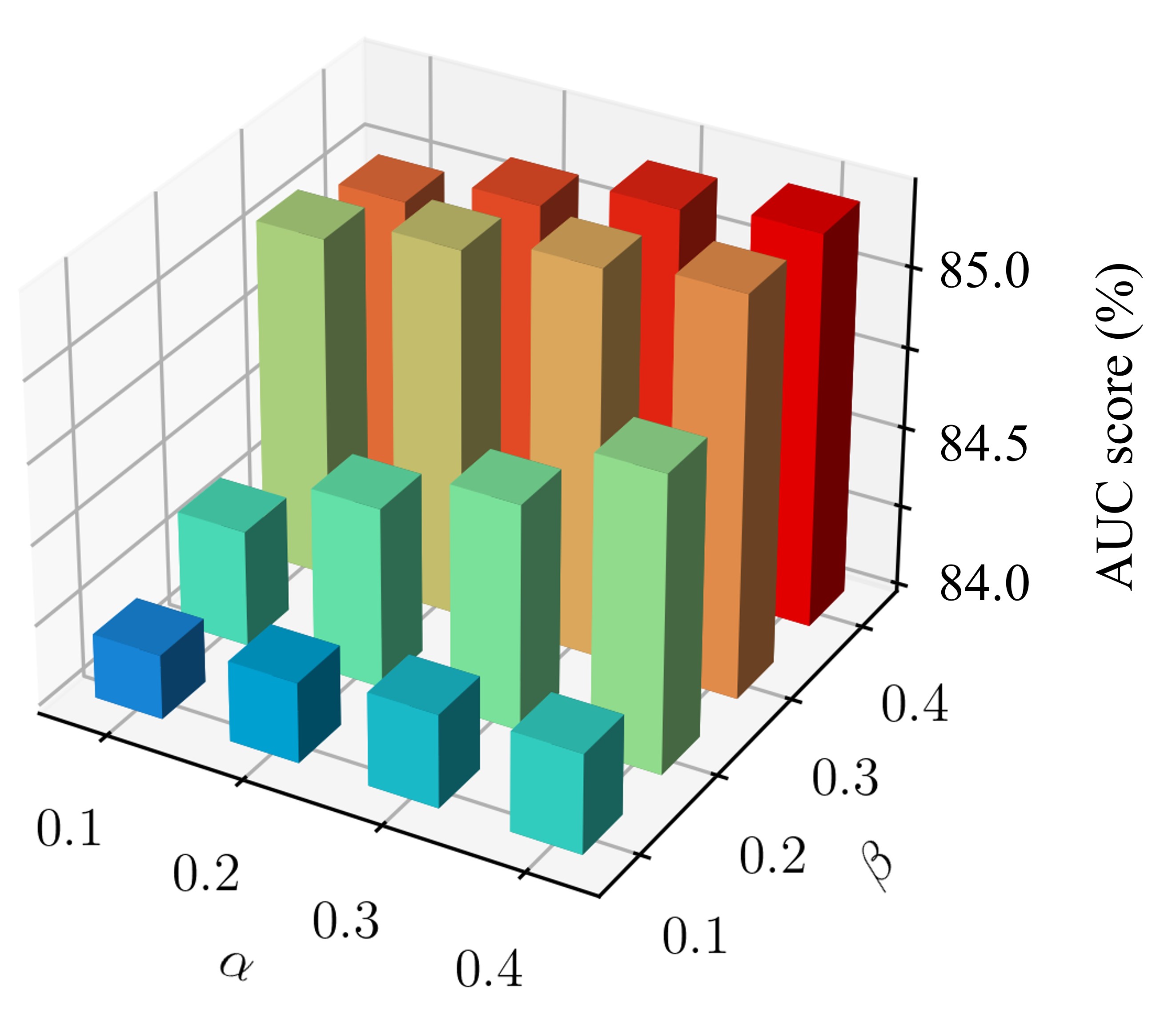}
    \caption{AUC score corresponding to different parameter combinations. The darker the color, the higher the corresponding AUC score.}
    \label{fig: parameters}
\end{figure}

\begin{figure}[htbp]
    \centering
    \includegraphics[width=6cm]{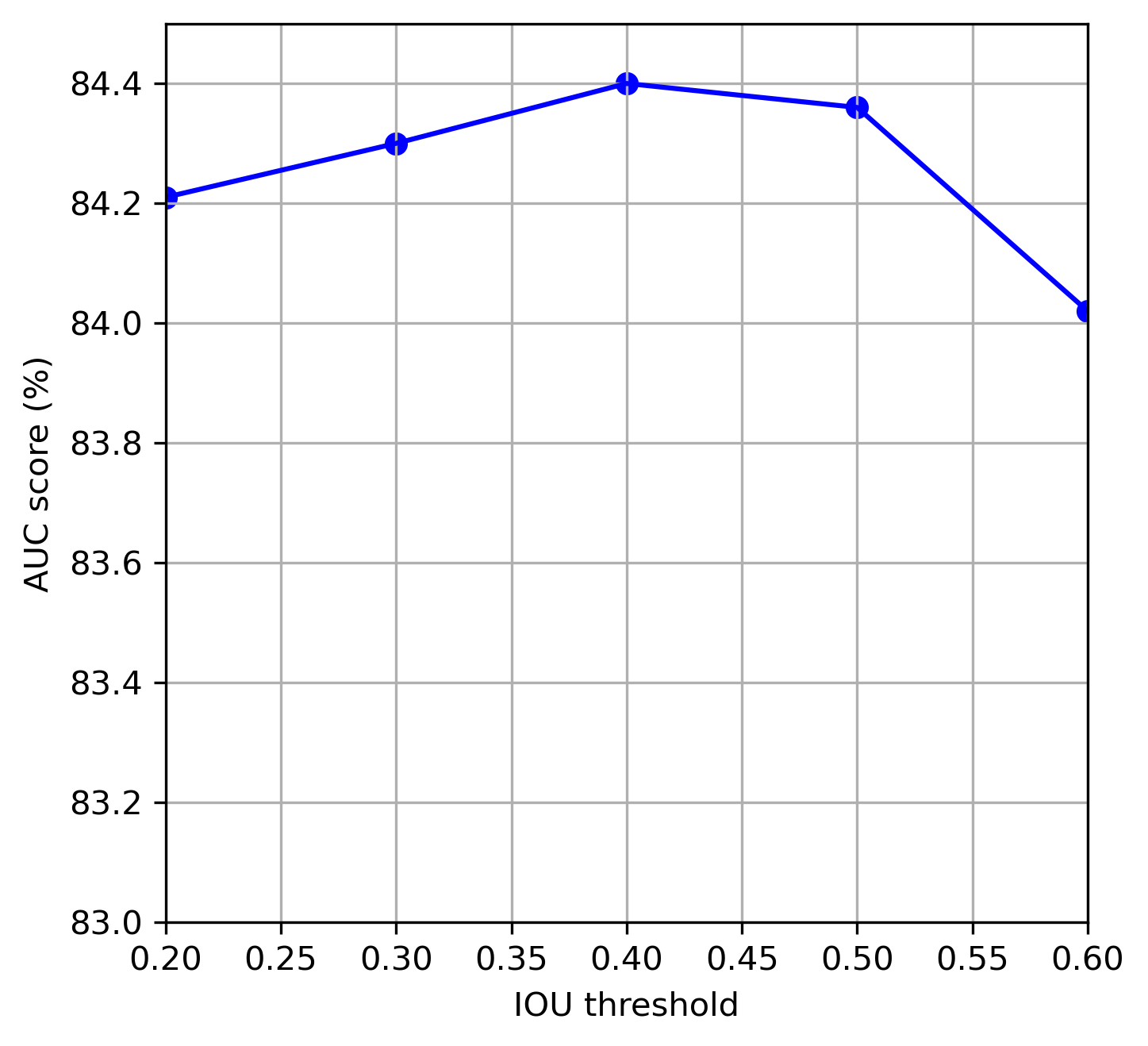}
    \caption{Relationship between IOU threshold and AUC score.}
    \label{fig: iou threshold}
\end{figure}

\subsection{Results}

\textbf{Hyper-parameter setting.}
    Then, we randomly split the dataset into training, validation, and test sets by the ratio of 8:1:1. We train on 0-1 labels and uncertain labels, respectively, and test on data that does not contain uncertain labels. In the feature extraction module, each image in the dataset is resized to $256 \times 256$. We normalized the image by mean ([0.485, 0.456, 0.406]) and standard deviation ([0.229, 0.224, 0.225]). The number of GCN network layers is set to 3 in graph network training. The ADAM optimizer is adopted with an empirical learning rate of 0.01 and momentum of 0.9. The batch size is set to 64 for training over 20 epochs. 

    In order to construct the spatial graph, we employed intersection over union (IOU) to measure the overlap between different anatomical areas and then set a threshold to determine whether to connect them. To investigate the impact of the IOU threshold on the spatial graph's performance, we experimented with different IOU thresholds, including 0.2, 0.3, 0.4, 0.5, and 0.6. Figure \ref{fig: iou threshold} showed that the performance of the spatial graph varied with different IOU thresholds.

    When the IOU threshold is set to a smaller value, such as 0.2, the AUC score is close to the result obtained by the implicit graph because almost all nodes are connected now. In contrast, only nodes with smaller distances are connected when the IOU threshold increases, reflecting the spatial graph's characteristics. As the IOU threshold becomes larger, there are fewer connections between nodes, which reduces the role of spatial information and ultimately results in a lower AUC performance. These findings suggest that the IOU threshold plays a crucial role in constructing an accurate spatial graph, and it needs to be carefully chosen depending on the specific task and data at hand.

    To find the best settings for the two parameters, we adjust the values of $\alpha$ and $\beta$ and record the AUC score for each parameter combination. Figure \ref{fig: parameters} reflects the performance of the AUC score corresponding to different parameter combinations. When the values of $\alpha$ and $\beta$ are small, the proportion of the spatial and semantic graphs is small, and the value of the AUC score is low. When the value of $\alpha$ and $\beta$ increase, the performance of the AUC score is improved. When the value of $\alpha$ and $\beta$ is around 0.3 and 0.4, the performance of the AUC score is the best.

\begin{figure*}[htbp]
    \centering
    \includegraphics[width=16cm]{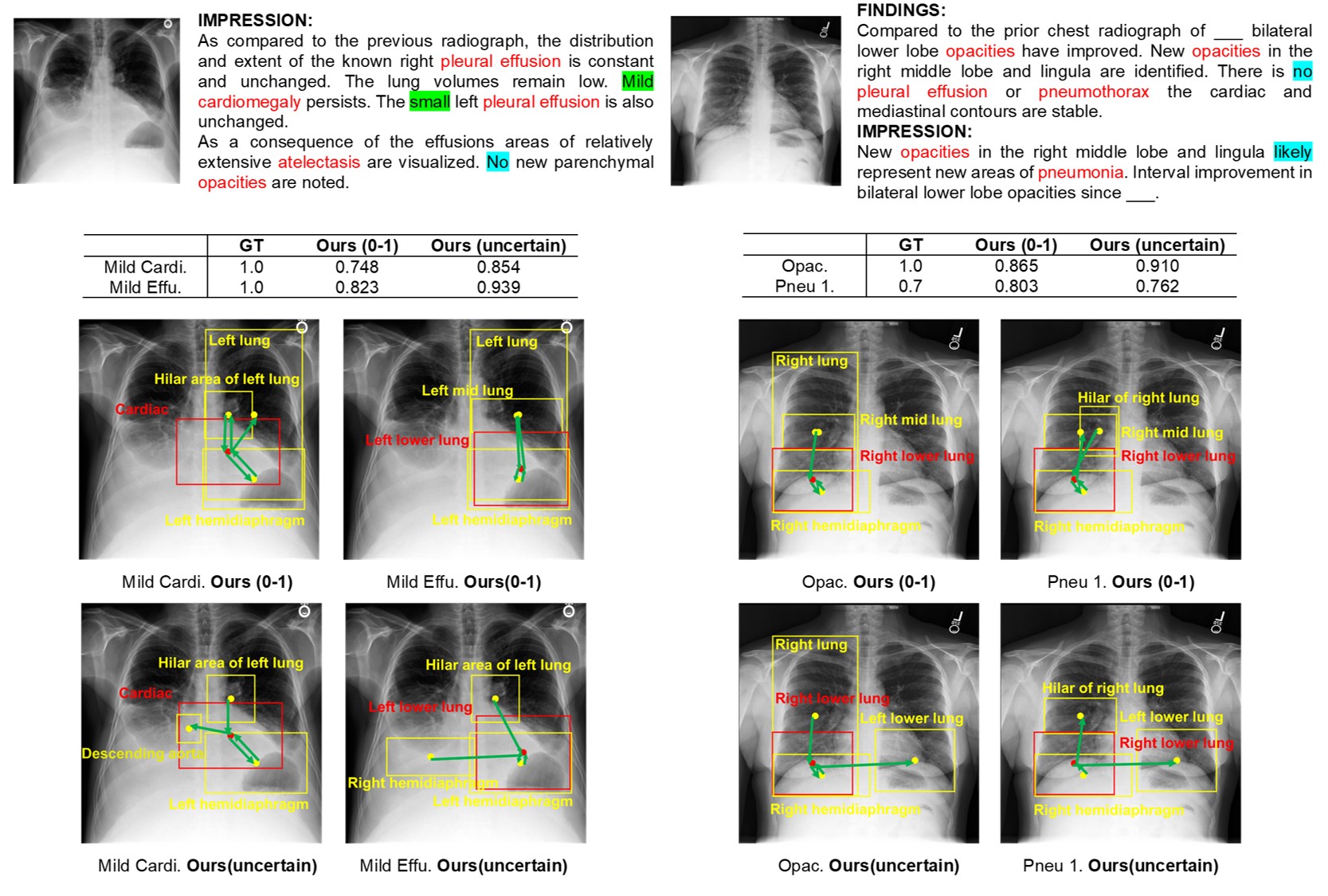}
    \caption{Results explanation. The left panel is a CXR image; the right is the corresponding report's findings and impression section. The table shows the results of training with 0-1 label and uncertain label on the graph model we built. The below images are the explanation of the given disease label. The red box indicates the most important node corresponding to the label. yellow boxes represent nodes that are closely related to the most important nodes. Each node corresponds to a box and represents the node's position at its center. We denote edges with normalized importance greater than 0.5 by green arrows.}
    \label{fig: explain}
\end{figure*} 

\vspace{0.3cm}
\textbf{Quantitative results.} 
    We calculated each method's mean AUC and top-k accuracy to measure the performance. In Table ~\ref{tab: experimental result}, when employing the ResNet-50 model as the backbone, training with uncertain labels can yield better performance than training with 0-1 labels. Compared with the ResNet-50 model, the graph model we built considers multiple relationships between anatomical structures and diseases, thus achieving better performance. The Sample Select model assigns a large weight to the clean label so that the network can remember the noise corresponding to the clean label, thereby improving the robustness of the model to the noisy label. Likewise, compared to training with 0-1 labels, utilizing uncertain labels on the graphical model we constructed yielded superior results, indicating the favorable properties of uncertain labels over 0-1 labels.

\vspace{0.3cm}
\textbf{Visualization of Example with Uncertainties Comparison.} 
    We present a series of examples showcasing the efficacy of several methodologies on test images, as illustrated in Figure \ref{fig: result}. The accompanying table lists the ground truth (GT) label obtained through our methodology and the performance metrics for different approaches. Categories with a substantial number of clean labels generally demonstrate superior classification performance, with several other methods exhibiting commendable results in these categories. On the other hand, our approach leverages uncertain labels extracted from reports, serving as a gold standard for supervision. As a result, our methodology outperforms other techniques in scenarios involving uncertain labels, yielding results closer to the actual labels.

\vspace{0.3cm}
\textbf{Comparison of visualization on interpretability with uncertainties.} 
    Interpretability methods are applied to our results to represent the underlying reasoning logic of the neural network’s decision. Figure \ref{fig: explain} explains given disease labels. It can be seen that the right effusion and left effusion are mentioned in the report. Medically speaking, effusion generally occurs at the bottom of the lung. Therefore, based on the judgment of medical knowledge, the positions of the important marked nodes cover the area where the disease occurs. Also, in cardiomegaly disease, important nodes calculated by our method cover the location of the heart. In addition, compared with the 0-1 label, the importance nodes found by the uncertain label are more consistent with the description in the report. For example, in the right effusion part, the nodes calculated by the 0-1 label do not cover the area at the bottom of the right lung. 

\section{Conclusion}

   In this study, using a rule-based approach, we developed a new dataset for classifying CXR images that incorporates expert uncertainty and severity-aware disease labels extracted from clinical notes. We proposed a novel multi-relationship graph learning technique using this dataset to predict expert-level uncertainties and disease severities. Our results suggest that this model can enhance the interpretability of existing medical imaging analysis frameworks, providing valuable insights into diagnosing and treating lung diseases. Our study contributes to developing more accurate and practical approaches for analyzing CXR images and improving patient care.

\section*{Acknowledgments}
This was supported in part by......

\bibliographystyle{unsrt}  
\bibliography{paper}

\begin{thebibliography}{10}

\bibitem{x.wangchestxray8hospitalscalechest2017}
{X. Wang}, {Y. Peng}, {L. Lu}, {Z. Lu}, {M. Bagheri}, and {R. M. Summers}.
\newblock {{ChestX-Ray8}}: {{Hospital-Scale Chest X-Ray Database}} and
  {{Benchmarks}} on {{Weakly-Supervised Classification}} and {{Localization}}
  of {{Common Thorax Diseases}}.
\newblock In {\em 2017 {{IEEE Conference}} on {{Computer Vision}} and {{Pattern
  Recognition}} ({{CVPR}})}, pages 3462--3471, July 2017.

\bibitem{irvinchexpertlargechest2019}
Jeremy Irvin, Pranav Rajpurkar, Michael Ko, Yifan Yu, Silviana {Ciurea-Ilcus},
  Chris Chute, Henrik Marklund, Behzad Haghgoo, Robyn Ball, Katie Shpanskaya,
  Jayne Seekins, David~A. Mong, Safwan~S. Halabi, Jesse~K. Sandberg, Ricky
  Jones, David~B. Larson, Curtis~P. Langlotz, Bhavik~N. Patel, Matthew~P.
  Lungren, and Andrew~Y. Ng.
\newblock {{CheXpert}}: {{A Large Chest Radiograph Dataset}} with {{Uncertainty
  Labels}} and {{Expert Comparison}}.
\newblock In {\em {{arXiv}}:1901.07031 [Cs, Eess]}, January 2019.

\bibitem{johnsonmimiccxrdeidentifiedpublicly2019}
Alistair E.~W. Johnson, Tom~J. Pollard, Seth~J. Berkowitz, Nathaniel~R.
  Greenbaum, Matthew~P. Lungren, Chih-ying Deng, Roger~G. Mark, and Steven
  Horng.
\newblock {{MIMIC-CXR}}, a de-identified publicly available database of chest
  radiographs with free-text reports.
\newblock {\em Scientific Data}, 6(1):317, December 2019.

\bibitem{x.ouyanglearninghierarchicalattention2021}
{X. Ouyang}, {S. Karanam}, {Z. Wu}, {T. Chen}, {J. Huo}, {X. S. Zhou}, {Q.
  Wang}, and {J. -Z. Cheng}.
\newblock Learning {{Hierarchical Attention}} for {{Weakly-Supervised Chest
  X-Ray Abnormality Localization}} and {{Diagnosis}}.
\newblock {\em IEEE Transactions on Medical Imaging}, 40(10):2698--2710,
  October 2021.

\bibitem{b.chenlabelcooccurrencelearning2020}
{B. Chen}, {J. Li}, {G. Lu}, {H. Yu}, and {D. Zhang}.
\newblock Label {{Co-Occurrence Learning With Graph Convolutional Networks}}
  for {{Multi-Label Chest X-Ray Image Classification}}.
\newblock {\em IEEE Journal of Biomedical and Health Informatics},
  24(8):2292--2302, August 2020.

\bibitem{yanglearnbeuncertain2019}
Hao-Yu Yang, Junling Yang, Yue Pan, Kunlin Cao, Qi~Song, Feng Gao, and Youbing
  Yin.
\newblock Learn to be uncertain: {{Leveraging}} uncertain labels in chest
  {{X-rays}} with bayesian neural networks.
\newblock In {\em Proceedings of the {{IEEE}}/{{CVF}} Conference on Computer
  Vision and Pattern Recognition ({{CVPR}}) Workshops}, June 2019.

\bibitem{zhoucontrastattentivethoracicdisease2021}
Yi~Zhou, Tianfei Zhou, Tao Zhou, Huazhu Fu, Jiacheng Liu, and Ling Shao.
\newblock Contrast-{{Attentive Thoracic Disease Recognition With Dual-Weighting
  Graph Reasoning}}.
\newblock {\em IEEE Transactions on Medical Imaging}, 40(4):1196--1206, April
  2021.

\bibitem{zhaocrosschestgraph2021}
Gangming Zhao.
\newblock Cross chest graph for disease diagnosis with structural relational
  reasoning.
\newblock In {\em Proceedings of the 29th {{ACM}} International Conference on
  Multimedia}, pages 612--620, {New York, NY, USA}, 2021. {Association for
  Computing Machinery}.

\bibitem{r.r.selvarajugradcamvisualexplanations22}
{R. R. Selvaraju}, {M. Cogswell}, {A. Das}, {R. Vedantam}, {D. Parikh}, and {D.
  Batra}.
\newblock Grad-{{CAM}}: {{Visual Explanations}} from {{Deep Networks}} via
  {{Gradient-Based Localization}}.
\newblock In {\em 2017 {{IEEE International Conference}} on {{Computer Vision}}
  ({{ICCV}})}, pages 618--626, 22.

\bibitem{rajpurkarchexnetradiologistlevelpneumonia2017}
Pranav Rajpurkar, Jeremy Irvin, Kaylie Zhu, Brandon Yang, Hershel Mehta, Tony
  Duan, Daisy Ding, Aarti Bagul, Curtis Langlotz, Katie Shpanskaya, Matthew~P.
  Lungren, and Andrew~Y. Ng.
\newblock {{CheXNet}}: {{Radiologist-Level Pneumonia Detection}} on {{Chest
  X-Rays}} with {{Deep Learning}}.
\newblock {\em arXiv:1711.05225 [cs, stat]}, December 2017.

\bibitem{z.lithoracicdiseaseidentification2018}
{Z. Li}, {C. Wang}, {M. Han}, {Y. Xue}, {W. Wei}, {L. -J. Li}, and {L.
  Fei-Fei}.
\newblock Thoracic {{Disease Identification}} and {{Localization}} with
  {{Limited Supervision}}.
\newblock In {\em 2018 {{IEEE}}/{{CVF Conference}} on {{Computer Vision}} and
  {{Pattern Recognition}}}, pages 8290--8299. {CVPR}, June 2018.

\bibitem{liualignattendlocate2019}
Jingyu Liu, Gangming Zhao, Yu~Fei, Ming Zhang, Yizhou Wang, and Yizhou Yu.
\newblock Align, {{Attend}} and {{Locate}}: {{Chest X-Ray Diagnosis}} via
  {{Contrast Induced Attention Network With Limited Supervision}}.
\newblock In {\em 2019 {{IEEE}}/{{CVF International Conference}} on {{Computer
  Vision}} ({{ICCV}})}, pages 10631--10640, {Seoul, Korea (South)}, October
  2019. {IEEE}.

\bibitem{g.zhaocontralaterallyenhancednetworks2021}
{G. Zhao}, {C. Fang}, {G. Li}, {L. Jiao}, and {Y. Yu}.
\newblock Contralaterally {{Enhanced Networks}} for {{Thoracic Disease
  Detection}}.
\newblock {\em IEEE Transactions on Medical Imaging}, 40(9):2428--2438,
  September 2021.

\bibitem{liknowledgedrivenencoderetrieve2019}
Christy~Y. Li, Xiaodan Liang, Zhiting Hu, and Eric~P. Xing.
\newblock Knowledge-{{Driven Encode}}, {{Retrieve}}, {{Paraphrase}} for
  {{Medical Image Report Generation}}.
\newblock {\em Proceedings of the AAAI Conference on Artificial Intelligence},
  33(01):6666--6673, July 2019.

\bibitem{wanglearningnoisylabels2021}
Deng-Bao Wang, Yong Wen, Lujia Pan, and Min-Ling Zhang.
\newblock Learning from {{Noisy Labels}} with {{Complementary Loss Functions}}.
\newblock {\em Proceedings of the AAAI Conference on Artificial Intelligence},
  35(11):10111--10119, May 2021.

\bibitem{renlearningreweightexamples2018}
Mengye Ren, Wenyuan Zeng, Bin Yang, and Raquel Urtasun.
\newblock Learning to {{Reweight Examples}} for {{Robust Deep Learning}}.
\newblock In {\em Proceedings of the 35th {{International Conference}} on
  {{Machine Learning}}}, pages 4334--4343. {PMLR}, July 2018.

\bibitem{y.lilearningnoisylabels22}
{Y. Li}, {J. Yang}, {Y. Song}, {L. Cao}, {J. Luo}, and {L. -J. Li}.
\newblock Learning from {{Noisy Labels}} with {{Distillation}}.
\newblock In {\em 2017 {{IEEE International Conference}} on {{Computer Vision}}
  ({{ICCV}})}, pages 1928--1936, 22.

\bibitem{huanguncertaintyawarelearninglabel2022}
Yingsong Huang, Bing Bai, Shengwei Zhao, Kun Bai, and Fei Wang.
\newblock Uncertainty-{{Aware Learning}} against {{Label Noise}} on
  {{Imbalanced Datasets}}.
\newblock In {\em Proceedings of the {{AAAI Conference}} on {{Artificial
  Intelligence}}}, volume~36, pages 6960--6969, June 2022.

\bibitem{xiasampleselectionuncertainty2021}
Xiaobo Xia, Tongliang Liu, Bo~Han, Mingming Gong, Jun Yu, Gang Niu, and Masashi
  Sugiyama.
\newblock Sample {{Selection}} with {{Uncertainty}} of {{Losses}} for
  {{Learning}} with {{Noisy Labels}}, June 2021.

\bibitem{petersonhumanuncertaintymakes2019}
Joshua Peterson, Ruairidh Battleday, Thomas Griffiths, and Olga Russakovsky.
\newblock Human {{Uncertainty Makes Classification More Robust}}.
\newblock In {\em 2019 {{IEEE}}/{{CVF International Conference}} on {{Computer
  Vision}} ({{ICCV}})}, pages 9616--9625, {Seoul, Korea (South)}, October 2019.
  {IEEE}.

\bibitem{zhanggeneralizedcrossentropy2018}
Zhilu Zhang and Mert~R. Sabuncu.
\newblock Generalized cross entropy loss for training deep neural networks with
  noisy labels.
\newblock In {\em Proceedings of the 32nd International Conference on Neural
  Information Processing Systems}, {{NIPS}}'18, pages 8792--8802, {Red Hook,
  NY, USA}, 2018. {Curran Associates Inc.}

\bibitem{lukasikdoeslabelsmoothing2020}
Michal Lukasik, Srinadh Bhojanapalli, Aditya~Krishna Menon, and Sanjiv Kumar.
\newblock Does label smoothing mitigate label noise?
\newblock In {\em Proceedings of the 37th International Conference on Machine
  Learning}, {{ICML}}'20. {JMLR.org}, 2020.

\bibitem{f.malearningnoisylabels2022}
{F. Ma}, {Y. Wu}, {X. Yu}, and {Y. Yang}.
\newblock Learning {{With Noisy Labels}} via {{Self-Reweighting From Class
  Centroids}}.
\newblock {\em IEEE Transactions on Neural Networks and Learning Systems},
  33(11):6275--6285, November 2022.

\bibitem{liuclassificationnoisylabels2016}
Tongliang Liu and Dacheng Tao.
\newblock Classification with {{Noisy Labels}} by {{Importance Reweighting}}.
\newblock {\em IEEE Transactions on Pattern Analysis and Machine Intelligence},
  38(3):447--461, March 2016.

\bibitem{hancoteachingrobusttraining2018}
Bo~Han, Quanming Yao, Xingrui Yu, Gang Niu, Miao Xu, Weihua Hu, Ivor~W. Tsang,
  and Masashi Sugiyama.
\newblock Co-teaching: {{Robust}} training of deep neural networks with
  extremely noisy labels.
\newblock In {\em Proceedings of the 32nd International Conference on Neural
  Information Processing Systems}, {{NIPS}}'18, pages 8536--8546, {Red Hook,
  NY, USA}, 2018. {Curran Associates Inc.}

\bibitem{jiangmentornetlearningdatadriven2018}
Lu~Jiang, Zhengyuan Zhou, Thomas Leung, Li-Jia Li, and Li~{Fei-Fei}.
\newblock {{MentorNet}}: {{Learning}} data-driven curriculum for very deep
  neural networks on corrupted labels.
\newblock In Jennifer Dy and Andreas Krause, editors, {\em Proceedings of the
  35th International Conference on Machine Learning}, volume~80 of {\em
  Proceedings of Machine Learning Research}, pages 2304--2313. {PMLR}, July
  2018.

\bibitem{hendrycksusingtrusteddata2018}
Dan Hendrycks, Mantas Mazeika, Duncan Wilson, and Kevin Gimpel.
\newblock Using trusted data to train deep networks on labels corrupted by
  severe noise.
\newblock In {\em Proceedings of the 32nd International Conference on Neural
  Information Processing Systems}, {{NIPS}}'18, pages 10477--10486, {Red Hook,
  NY, USA}, 2018. {Curran Associates Inc.}

\bibitem{hanmaskingnewperspective2018}
Bo~Han, Jiangchao Yao, Gang Niu, Mingyuan Zhou, Ivor Tsang, Ya~Zhang, and
  Masashi Sugiyama.
\newblock Masking: {{A}} new perspective of noisy supervision.
\newblock In S.~Bengio, H.~Wallach, H.~Larochelle, K.~Grauman,
  N.~{Cesa-Bianchi}, and R.~Garnett, editors, {\em Advances in Neural
  Information Processing Systems}, volume~31. {Curran Associates, Inc.}, 2018.

\bibitem{renfasterrcnnrealtime2015}
Shaoqing Ren, Kaiming He, Ross Girshick, and Jian Sun.
\newblock Faster {{R-CNN}}: {{Towards Real-Time Object Detection}} with
  {{Region Proposal Networks}}.
\newblock In {\em Advances in {{Neural Information Processing Systems}}},
  volume~28. {Curran Associates, Inc.}, 2015.

\bibitem{wuchestimagenomedataset2021}
Joy~T Wu, Nkechinyere Agu, Ismini Lourentzou, Ismini Lourentzou, Arjun Sharma,
  Joseph~Alexander Paguio, Jasper~Seth Yao, Edward~C Dee, William Mitchell,
  Satyananda Kashyap, Andrea Giovannini, Leo~Anthony Celi, and Mehdi Moradi.
\newblock Chest {{ImaGenome Dataset}} for {{Clinical Reasoning}}.
\newblock In J.~Vanschoren and S.~Yeung, editors, {\em Proceedings of the
  {{Neural Information Processing Systems Track}} on {{Datasets}} and
  {{Benchmarks}}}, volume~1, 2021.

\bibitem{r.girshickfastrcnn07a}
{R. Girshick}.
\newblock Fast {{R-CNN}}.
\newblock In {\em 2015 {{IEEE International Conference}} on {{Computer Vision}}
  ({{ICCV}})}, pages 1440--1448, 7.

\bibitem{zhangwhenradiologyreport2020}
Yixiao Zhang, Xiaosong Wang, Ziyue Xu, Qihang Yu, Alan Yuille, and Daguang Xu.
\newblock When {{Radiology Report Generation Meets Knowledge Graph}}.
\newblock In {\em Proceedings of the {{AAAI Conference}} on {{Artificial
  Intelligence}}}. {AAAI}, February 2020.

\bibitem{popeexplainabilitymethodsgraph2019}
Phillip~E. Pope, Soheil Kolouri, Mohammad Rostami, Charles~E. Martin, and Heiko
  Hoffmann.
\newblock Explainability {{Methods}} for {{Graph Convolutional Neural
  Networks}}.
\newblock In {\em 2019 {{IEEE}}/{{CVF Conference}} on {{Computer Vision}} and
  {{Pattern Recognition}} ({{CVPR}})}, pages 10764--10773, {Long Beach, CA,
  USA}, June 2019. {IEEE}.

\end{thebibliography}

\end{document}